\documentclass{article}

\usepackage{PRIMEarxiv}

\usepackage[utf8]{inputenc} 
\usepackage[T1]{fontenc}    
\usepackage{hyperref}       
\usepackage{url}            
\usepackage{booktabs}       
\usepackage{amsfonts}       
\usepackage{nicefrac}       
\usepackage{microtype}      
\usepackage{lipsum}
\usepackage{caption}
\usepackage{booktabs}
\usepackage{multirow}
\usepackage{subcaption}
\usepackage{fancyhdr}       
\usepackage{graphicx}       
\graphicspath{{media/}}     

\title{High-Level Context Representation for Emotion Recognition in Images}

\author{
  Willams de Lima Costa\\
  Voxar Labs, Centro de Informática\\
  Universidade Federal de Pernambuco\\
  Recife, Brazil\\
  \texttt{wlc2@cin.ufpe.br} \\
   \And
  Estefania Talavera Martinez \\
  Data Management and Biometrics Group \\
  University of Twente \\
  Enschede, The Netherlands\\
  \And
  Lucas Silva Figueiredo \\
  Unidade Acadêmica de Belo Jardim \\
  Universidade Federal Rural de Pernambuco \\
  Belo Jardim, Brazil \\
  \And
  Veronica Teichrieb \\
  Voxar Labs, Centro de Informática \\
  Universidade Federal de Pernambuco \\
  Recife, Brazil \\
}

\begin{document}
\maketitle

\begin{abstract}
Emotion recognition is the task of classifying perceived emotions in people. Previous works have utilized various nonverbal cues to extract features from images and correlate them to emotions. Of these cues, situational context is particularly crucial in emotion perception since it can directly influence the emotion of a person. In this paper, we propose an approach for high-level context representation extraction from images. The model relies on a single cue and a single encoding stream to correlate this representation with emotions. Our model competes with the state-of-the-art, achieving an mAP of 0.3002 on the EMOTIC dataset while also being capable of execution on consumer-grade hardware at $\approx 90$ frames per second. Overall, our approach is more efficient than previous models and can be easily deployed to address real-world problems related to emotion recognition.
\end{abstract}

\section{Introduction}
\label{sec:intro}

Among diverse possible representations of human behavior, emotion recognition is a research topic that gained interest in the last few years, mainly due to the high development of application scenarios that can explore affective computing concepts, such as smart environments and social robotics. By understanding not only the user but also the person, systems can act on top of behavioral knowledge and propose diverse interventions to improve interaction, experience, or even quality of life.

To recognize emotions, systems must first decode emotional signals sent naturally by humans, sometimes even without intention \cite{buck88}. These signals are usually referred to as \textit{implicit} and are derived from a form of communication defined as \textit{nonverbal communication}. Research shows that a significant amount of affective information is not communicated verbally but is sent and received naturally through this communication channel, which encompasses facial expressions, body language, speech tonality, and other emotional cues \cite{rouast, patel2014body}. This ability allows for an unobtrusive experience and a perception of emotion that is not posed.

However, in unrestricted in-the-wild scenarios, our emotions are also influenced by information in the situational context. For example, a person sitting on a beach, enjoying the sun and the sea during their vacation, is more likely to experience positive sentiments such as joy and happiness. In contrast, a person stuck in a traffic jam filled with noise pollution could be inclined towards more negative sentiments such as frustration, anger, or stress. Therefore, environmental stimuli should also be taken into consideration when analyzing emotion.

Researchers have been proposing approaches that take into consideration contextual information for a while now in works such as EMOTIC \cite{kosti2017emotion}, CAER-Net \cite{caer}, EmotiCon \cite{Mittal_2020_CVPR} GLAMOR-Net \cite{glamor}, and EmotionRAM \cite{costa4255748fast}, each proposing new approaches on how to leverage context and extract its representations from images on different datasets. We hypothesize that, although working on these low-level representational features could and has led to significant results in the past, generating semantic, high-level descriptions could be more assertive to unseen data, leading to better results in test sets and when deployed to solve real-world problems.

In some scenarios, high-level representations of emotions can be a valuable aid for decision-makers to make informed choices. For example, consider a city planner that needs to decide which public parks in the city need to be improved or renovated first. To make this decision, the planner needs to know how people feel when they are in these spaces, but they may not need to know each individual experience to make this decision. Instead, the high-level representation could be provided as an overview of the emotions associated with that context and how people act towards it. It could be easily compared without needing to act on top of a significant amount of data. Approaches such as this one have several advantages, such as resource-saving.

In this work, we propose an approach for the extraction of high-level context representations of images for the task of emotion recognition. We show in our experiments that these highly-representative descriptions of context are capable of yielding results comparable to the state-of-the-art of emotion recognition on the EMOTIC dataset \cite{kosti2019context} by itself and could easily be placed into a complete emotion recognition pipeline as a context encoding module to lead to significant improvements in accuracy. We also show that our proposal can perform a fast inference, a desirable feature for low-consumption edge devices that could be deployed in the wild. The contributions of this work are as follows:

\begin{itemize}
    \item We propose a novel framework that (i) builds a high-level representation of extracted context descriptors from images and (ii) employs Graph Convolutional Networks (GCNs) to classify these representations into emotional categories.
    \item We benchmark on the well-known EMOTIC dataset \cite{kosti2019context}, achieving a comparable accuracy with the state-of-the-art.
    \item We discuss how our model's low computational power requirements make diverse applications possible to solve real-world problems.
\end{itemize}

The rest of the paper is organized as follows: Section \ref{section:soa} provides an overview of works for emotion recognition from images. Our proposed framework for high-level context description is described in Section \ref{section:method}. Later, in Sections \ref{sec:exp} and \ref{sec:res_dis}, we depict our experimental setup and discuss the obtained results, respectively. Finally, we draw conclusions in Section \ref{sec:concl}

\section{Literature review}
\label{section:soa}

It is well-known that facial expressions contain very descriptive features related to emotion. According to Barret, faces are to us as words are on a page, and humans can decode them perceptually with some level of ease \cite{barrett2007language}. Naturally, researchers have been investigating how to extract these features in a task called Facial Expression Recognition (FER). However, some challenges come when evaluating FER in uncontrolled scenarios. The face is, many times, partially or fully occluded, and techniques that depend solely on this information might not be able to perform well.

Extending these works, researchers have been investigating how adding other nonverbal cues could improve the pipeline. Kosti \textit{et al.} \cite{kosti2017emotion} proposed a baseline for this approach, in which both the person and the context in which they are placed would be considered. Therefore, in the case of occlusion, there would be other contributors to perceived emotion. They also published the EMOTIC dataset that serves as a baseline for validating these types of approaches. EMOTIC is still used widely in the literature to validate techniques based on context.

Other techniques have also been proposed, such as CAER-Net \cite{caer} and GLAMOR-Net \cite{glamor}, which employ different formats of how to weigh context contributions and, therefore, how important should be the contextual information in each scenario. However, all of these techniques mentioned above have the same limitation by design: the lack of definition of what should be considered as context. For example, the approach proposed by \cite{glamor} considers detecting the face of a person, completely occluding it with a black rectangle, and using this new image as a representation of context. However, the other body parts are still visible, as are the body parts of other people in the scene, and this image would be fed to a context encoding stream that is designed to extract features from the scene automatically. However, are these encoding streams capable of doing this task without prior knowledge?

Other approaches, such as EmotiCon \cite{Mittal_2020_CVPR}, proposed that it is necessary to use multiple independent and specialist streams to generate representations that can be correlated to emotion, given how context is highly descriptive. Specifically for EmotiCon, as an example, the authors propose the usage of the following context streams: (1) multimodal context, with facial landmarks and body keypoints; (2) situational context, extracted by processing the background image with the person occluded by using a pedestrian tracking method; and (3) socio-dynamic context, which computes proximity features using depth maps.

In a more recent approach, Chen \textit{et al.} \cite{chen2023incorporating} proposed models combining different representations from context. For example, they use a deep network for each person on the scene to calculate their social relations between intimate, not intimate, and no relation. They also propose a deep reasoning module for using multiple context representations that are extracted locally and globally and involve scene recognition and body pose estimation, among other modules.

\begin{figure*}[t!]
\centering
\begin{subfigure}{0.33\textwidth}
\centering
\includegraphics[width=\columnwidth]{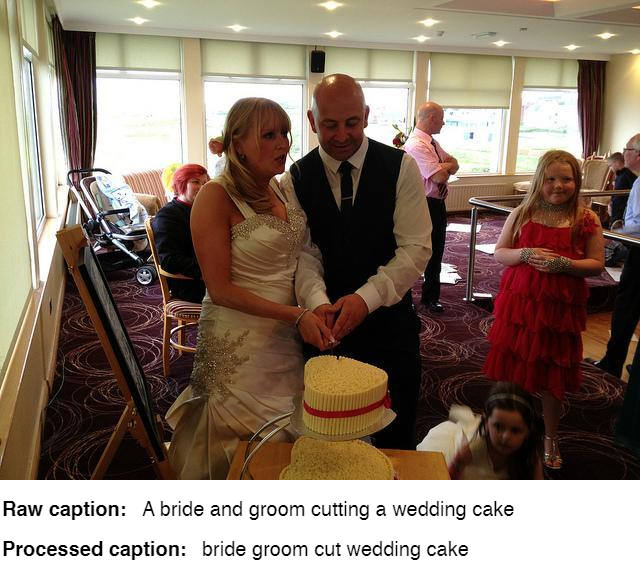}
\caption{}
\label{fig:captions_a}
\end{subfigure}
\begin{subfigure}{0.33\textwidth}
\centering
\includegraphics[width=\columnwidth]{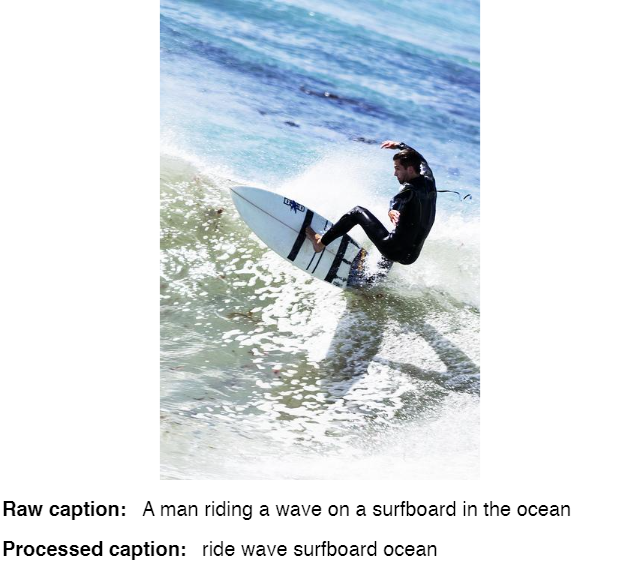}
\caption{}
\label{fig:captions_b}
\end{subfigure}
\begin{subfigure}{0.33\textwidth}
\centering
\includegraphics[width=\columnwidth]{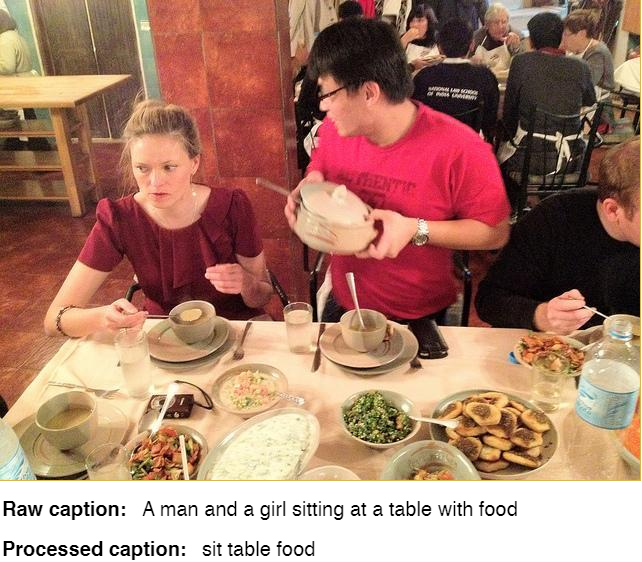}
\caption{}
\label{fig:captions_c}
\end{subfigure}

\caption{Examples of images from EMOTIC with their raw captions and processed captions.}
	\label{fig:captions}
\end{figure*}

However, humans perceive context differently \cite{barrett2010context, barrett2011context}. The literature suggests that humans encode context naturally by using our internal representations of meaning in the image. Therefore, it is not natural for us to take calculated steps to understand context. Instead, our brain automatically classifies these stimuli as positive, neutral, or negative based on our previous knowledge of that information \cite{pastor2008affective}. Our approach differs from the techniques mentioned earlier due to the more straightforward approach for context, in which we try to mimic the context representation of humans based on our best knowledge of the literature on nonverbal communication and behavioral psychology.

\section{High-level context representation}
\label{section:method}
In this section, we describe our approach for extracting high-level representations from context. Given how humans describe and understand context in images, we propose extracting high-level descriptions of images to correlate them with semantic features. This approach mimics how humans correlate semantic descriptions with emotions to improve interpretability.

\subsection{Extraction of image descriptors}
\paragraph{High-level descriptions.} Given an image as input, we first want to extract high-level image descriptions. We employ ExpansionNet-v2 \cite{hu2022expansionnet}, an image captioning model based on the Swin-Transformer architecture \cite{liu2021swin}. We first traverse through the EMOTIC dataset, and for each sample, we input the image to ExpansionNet-v2 for captioning generation.

We then process the raw caption to generate a refined caption. First, we perform the removal of stop words from the caption. Stop words are common words in a language, such as articles, prepositions, and pronouns, but do not have any semantic meaning. Therefore, maintaining these words would only elevate the complexity, given their high frequency in the English language, and by removing them, we have a more representative corpus. We use spaCy\footnote{Available at \url{https://spacy.io/}}, a public library for natural language processing. We also remove common nouns such as \textit{man}, \textit{woman}, \textit{girl}, and \textit{boy}. We also applied lemmatization to reduce each word to its root form. The remaining words are called \textit{valid words} and will be used in the following steps to generate data representations. Finally, we show examples of images, their original captions, and their processed captions in \autoref{fig:captions}.

\paragraph{Co-occurrence mining.} The second step involves the generation of co-occurrence matrices that will represent patterns of labels within the dataset, which will be employed in the future through conditional probability. After preprocessing the captions in the dataset, we store this information and count the occurrence of each emotion, and the valid words of each caption, resulting in a matrix \begin{math}M_c \in \mathbb{N}^{ W \times C }\end{math}, where \(W\) is the number of valid words from the corpus and \(C\) is the number of emotion categories in the dataset. Therefore, \(M_{c_{ij}}\) denotes the number of times that emotion \(C_j\) occurred when the valid word \(W_i\) also occurred. We call this matrix the emotion co-occurrence matrix.
\begin{figure*}[t!]
	\centering
    \includegraphics[width=\textwidth]{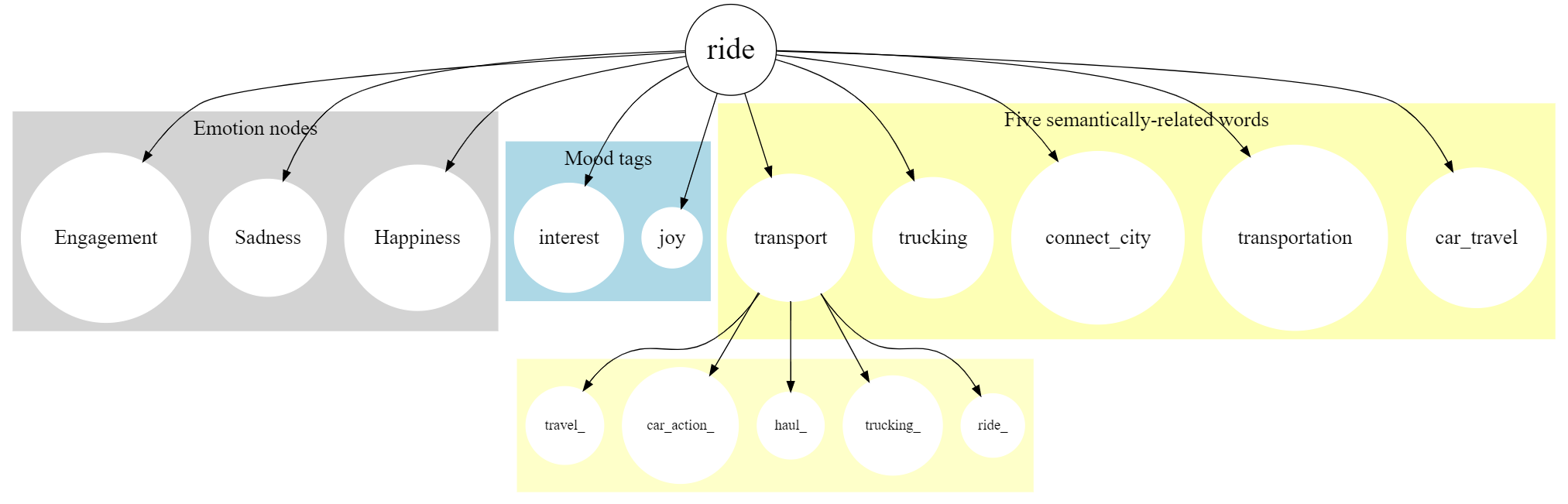}
	\caption{Example of the generated graph. For brevity and visualization, we consider only one valid word in this scenario. For each valid word node, we create nodes for emotions (also reduced for brevity), mood tags, and semantically-related words. For each semantically-related word, we also query SenticNet and extract their semantically-related words, as shown in the "transport" node.}
	\label{fig:graph}
\end{figure*}
Based on the same assumption, we also generate a co-occurrence matrix based on the co-occurrence of valid words. Given a window of size \(s\), we slide this window to capture the co-occurrence of the valid words, resulting in a matrix \begin{math}M_w \in \mathbb{N}^{ W \times W}\end{math}. Therefore, \(M_{W_{ij}}\) denotes the number of times that the valid word \(W_i\) appeared together with the valid word \(W_j\).

\paragraph{Semantic descriptions.} For each valid word \(W\), we extract semantic representations that can be correlated with emotion. Given how ExpansionNet-v2 is a model to generate captions in a generic context, extracting the semantic representations of the word will lead to better representations of affective meaning in that caption.

For extracting semantic descriptions, we employ SenticNet \cite{cambria2015senticnet}, a knowledge base for semantics, sentics, and polarity associated with natural language concepts. We query each valid word, and we extract the following attributes: the two mood tags associated with the concept; the pleasantness sentic, which represents the perception of pleasantness or unpleasantness of the word; the polarity value, which represents the overall sentiment of the word and finally, the semantically-related concepts.

Except for the valid word not existing on the SenticNet knowledge base, we use WordNet \cite{miller1995wordnet, miller1998wordnet} to search for synonyms. The advantage of this approach is that the words in WordNet are grouped using synsets, which are sets of synonyms with similar concepts or meanings. Therefore, by querying a word in WordNet, its synonym will have a significant relationship and most surely have the same meaning. For each possible synonym, we rank the list according to the similarity with the valid word and iterate through the list, selecting the first synonym present in SenticNet. In the rare case that the valid word is not present on SenticNet and neither are its synonyms, we drop the valid word from the caption and proceed to the next step without it.

\paragraph{Graph generation.} With this prior knowledge (e.g., co-occurrence and semantic representations), we can capture relationships between valid words and emotions and also between themselves. Given how they are a particularly effective method of describing structured data, we choose to model these representations using graphs. Although some of the knowledge is learned prior, the definition and construction of graphs are done as needed and in real-time. This allows this technique to generate representations from unseen data.

We use Deep Graph Library\footnote{Available at \url{www.dgl.ai}}, a framework-agnostic library for generating and manipulating graphs. We start by constructing an empty graph \(G = (V, E)\), in which \(V\) is a set of nodes and \(E\) is a set of edges. In this case, \(V = E = \{\emptyset\}\). For each valid word \(W\), we start by adding a new node \(V_{W_{i}}\) to the before empty set of nodes \(V\) of the graph. We use GloVe \cite{pennington2014glove} to fetch the valid word embedding and use this representation as the feature \(X \in \mathbb{R}^{50}\) for node \(V_{W_i}\). If the valid word is absent on GloVe, we randomly sample this embedding from a uniform distribution [-0.01, 0.01]. We save this representation for future use in case this valid word reappears.

Next, we add a node \(V_{C}\) for each emotion category \(C\) in the dataset. For EMOTIC, since we have 26 possible emotions, we add 26 nodes and place edges \(e = (V_W, V_{C_{i}})\) between the valid word and each emotion. We define the weight \(w_e\) according to the equation below:

\begin{equation}
    w_e = P(C_i|W) = \frac{M_{C_{W,i}}}{\sum M_{C_{W}}},
\end{equation}

where the edge weight \(w_e\) between the valid word node \(V_W\) and the \(i_{th}\) emotional category \(C_i\) is \(P(C_i|W)\), which is given by the co-occurrence between the valid word \(W\) and emotion category \(C_i\) divided by the sum of the co-occurrence between the valid word \(W\) and all possible emotions \(C\), extracted from the co-occurrence matrix \(M\).

Next, we add nodes related to the sentic semantic description of the word. First, we add two nodes relative to the mood tags extracted from SenticNet, and we set the weights of the edge between the valid word node and them to be the pleasantness value of the word. Next, we add five nodes relative to the five semantically-related words available from querying SenticNet and add edges using the polarity value as weight.

For each of the five semantically-related words to the valid word, we also query SenticNet and extract their five semantically-related words. The polarity value of the word in the first level gives the edge between these connections. We hypothesize that by adding another level of semantic relationships, we will be able to extract even deeper representations of context.

We perform the same process for each valid word \(W\) in the caption. After the nodes for all valid words are created, we add edges between these nodes. The weight of each edge is given by the co-occurrence of the words \(M_{W_{i, j}}\), divided by the total number of times the word appeared. Finally, we show an example of the graph with reduced information for brevity in \autoref{fig:graph}.

\subsection{Deep GCN for Emotion Recognition} 

Given the construction of graphs to represent context, we use a deep graph convolutional neural network for graph classification and, consequently, for emotion recognition. Given a set of graphs \({G_1, ..., G_N}\) and a set of emotion categories \(C \in \mathbb{R}^{26}\), we aim to classify each graph according to an emotional category. For this task, we propose adapting Graph Isomorphism Network (GIN) \cite{xu2018powerful}, chosen due to its simple architecture, which could lead to reasonable inference rates in low-energy, low-consumption devices.

First, given a graph as input, we store this graph's features directly in the hidden representations stack as \(h0\). After this, we loop through a GIN convolutional block containing a GIN layer, batch normalization, and ReLU. We iterate over this block five times in this approach, generating representations \(h1\) to \(h5\). Finally, we iterate through the hidden representations, average pooling these features and reducing their dimensionality. In parallel, we keep a stream for the categorical classification, which outputs classification labels \(C\), and another stream for continuous predictions for a VAD model. The VAD model \cite{vad1} is a common approach for emotion recognition which represents emotions in a three-dimensional space, where each dimension corresponds to a different aspect of the emotional experience. 

The VAD model is composed of Valence, Arousal, and Dominance. The Valence (V) axis refers to how pleasant or unpleasant emotion is to the individual experiencing it. Arousal (A) represents the level of energy associated with the emotion; high arousal, for example, is related to more energetic, excited emotions, while lower arousal is to more calm emotions. Finally, Dominance (D) reflects the degree of control or power the person feels over this emotional state. Researchers have found that different emotions tend to cluster in regions of the VAD space, reflecting their shared aspects and underlying representation. Therefore, each emotion can be represented as a linear combination of the three axes \cite{vad2}.

Therefore, we learn categorical labels and continuous values during training. We define our loss as a weighted combination of the individual losses of each output. Given a prediction \begin{math}\hat{y} = (\hat{y}_{cat}, \hat{y}_{cont})\end{math} in which \begin{math}\hat{y}_{cat} \in \mathbb{R}^C\end{math} and \begin{math}\hat{y}_{cont} \in \mathbb{R}^3\end{math}, we define the loss in this prediction as \begin{math}L = \lambda_{cat}L_{cat} + \lambda_{cont}L_{cont}\end{math}, where \(L_{cat}\) and \(L_{cont}\) represents the loss of each individual prediction. For \(L_{cat}\) implement a weighted euclidean loss as used in EMOTIC \cite{kosti2019context}, which is defined as follows:

\begin{equation}
L_{2_{cat}}(\hat{y}_{cat}) = \sum_{i=1}^{26} w_i(\hat{y}_{cat_i}-y_{cat_i})^2,
\end{equation}

\noindent in which \(\hat{y}_{cat_i}\) is the prediction for the \(i_{th}\) category and \({y}_{cat_i}\) is its ground-truth label. The weight \(w_i\) is defined as \(w_i = \frac{1}{ln(c+p_i)}\), where \(p_i\) is the probability of the \(i_{th}\) category and \(c\) is a parameter to control the range of valid values. We also employ a L2 loss for \(L_{cont}\), defined as:

\begin{equation}
L_{2_{cont}}(\hat{y}_{cont}) = \sum_{j=1}^{3} (\hat{y}_{cont_j}-y_{cont_j})^2.
\end{equation}

\begin{figure*}[t!]
	\centering
	\includegraphics[width=\textwidth]{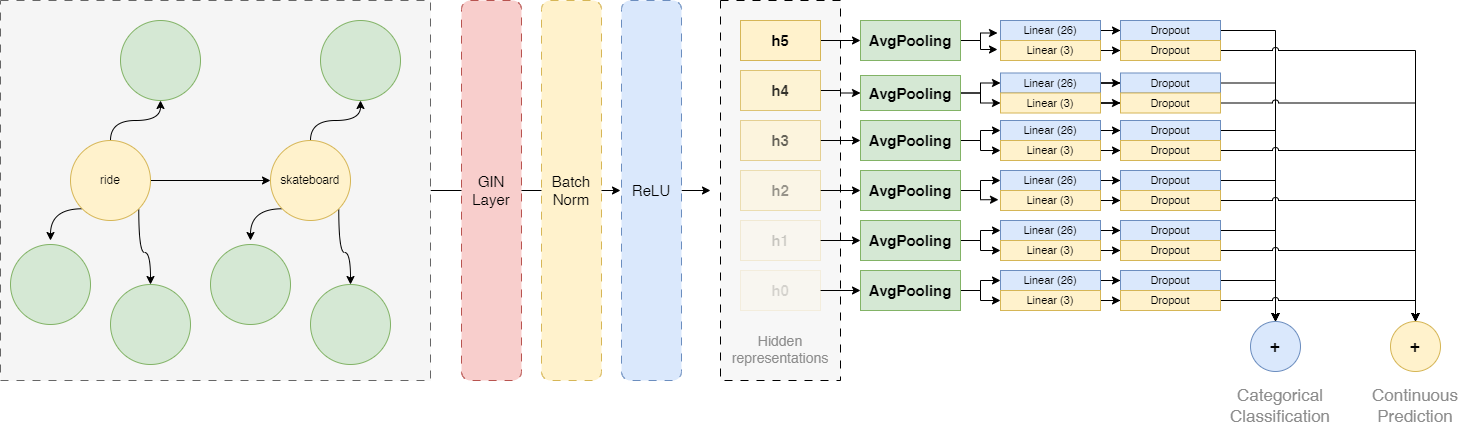}
	\caption{Our proposed architecture for high-level context representation. Given an input graph that is generated using previously learned knowledge and image captioning, we use an adaptation of Graph Isomorphism Network (GIN) \cite{xu2018powerful} to classify it among a set of emotions. Given how EMOTIC also has annotations using a continuous model, we adapt the pipeline to generate two predictions, which are considered to calculate the loss.}
	\label{fig:model}
\end{figure*}

Finally, we train our model on the EMOTIC dataset using the abovementioned features. We use the default PyTorch data loader and implement access to the list of graphs to feed the model during execution. We show an overview of our model on \autoref{fig:model}.

\section{Experiments}
\label{sec:exp}
\paragraph{Dataset.} We perform our experiments on the Emotions in Context dataset (EMOTIC) \cite{kosti2019context}, using the 2019 version, to allow direct comparison with the state-of-the-art. This dataset is focused on samples of people in their unposed environment and, therefore, comprises images with real, felt emotions and expressions. Another focus of this dataset is to have images with context visibility and variability, meaning that the images are framed with contextual information and that various contexts are present among the samples. The authors also gathered samples from other well-established datasets, such as COCO \cite{lin2014microsoft} and ADE20K \cite{zhou2019semantic}, extending these images with emotion annotations.

EMOTIC was annotated using the Amazon Mechanical Turk service. The images on the training set were annotated only once, while the test and validation sets were annotated by three and five annotators, respectively. According to the authors, the dataset contains 23,571 images with 34,420 annotated subjects, 66\% males and 34\% females, 10\% children, 7\% teenagers, and 83\% adults. Differently from what the literature on behavioral psychology suggests \cite{ekman92basic}, EMOTIC employs 26 emotional categories proposed based on word connections and inter-dependence of words, forming, therefore, word groupings. However, the dataset lacks the \textit{Neutral} category, which can harm the annotation process.

\paragraph{Comparison with the state-of-the-art.} We compare our results with other techniques of state-of-the-art, namely EMOTIC\cite{kosti2019context}, Zhang's work \cite{zhang2019context}, EmotiCon \cite{Mittal_2020_CVPR}, DRM \cite{chen2023incorporating}, LEKG \cite{chen2023incorporating}, which are two variations of Chen's method \cite{chen2023incorporating}, and Yang's work \cite{yang2022emotion}. However, as we later describe in Table \ref{sota}, these techniques are often built on top of a combination of multiple nonverbal cues. 

\paragraph{Validation metrics.} Besides the quantitative evaluation using the mean average precision (mAP) metric, as is done in the current literature \cite{kosti2017emotion, kosti2019context, zhang2019context, Mittal_2020_CVPR, chen2023incorporating}, we present some examples to perform a brief qualitative evaluation of the predictions.

\paragraph{Implementation details.} We train our model from scratch, learning the parameters using Adadelta \cite{zeiler2012adadelta}. After an empirical comparison of multiple values on the validation set, the batch size is set to \(16\). We use a learning rate of \(0.001\) and a weight decay of \(
0.0004\).

Regarding the experimentation environment, we train and validate our model on a desktop computer running Ubuntu 20.04 LTS with an Intel i7-4790K with 32 GB of RAM and an NVIDIA RTX 2080 Ti with 12GB of VRAM. For training and experimenting with our model, we use PyTorch 1.12 with CUDA 11.3 and CuDNN 8.3.2. For the experimentations regarding inference time, we also compare it with a consumer-grade notebook with Windows 10 Pro, 16 GB of RAM, and an NVIDIA GeForce GTX 1060M with 6GB of VRAM.
\begin{table}[t!]
\caption{Quantitative evaluation of our approach compared with state-of-the-art models on EMOTIC dataset.}
\centering
\resizebox{0.5\columnwidth}{!}{%
\renewcommand{\arraystretch}{1.1}
\begin{tabular}{lll} 
\toprule
\textbf{Technique} & \textbf{mAP} & \textbf{\# of nonverbal cues} \\ 
\hline\hline
\multirow{2}{*}{EMOTIC \cite{kosti2019context}} & 27.38 & \multirow{2}{*}{2 (body and context)} \\
 & 20.47* &  \\ 
\midrule
Zhang \cite{zhang2019context} & 28.42 & 1 (context with two streams) \\ 
\midrule
\multirow{2}{*}{EmotiCon \cite{Mittal_2020_CVPR}} & 35.48 & \multirow{2}{*}{\begin{tabular}[c]{@{}l@{}}4 (face description, body pose, context\\and depth mapping)\end{tabular}} \\
 & 26.87* &  \\ 
\midrule
DRM \cite{chen2023incorporating} & 26.48 & \begin{tabular}[c]{@{}l@{}}5 (three body descriptions and two\\context descriptions)\end{tabular} \\ 
\midrule
LEKG \cite{chen2023incorporating} & 29.47 & \begin{tabular}[c]{@{}l@{}}2 (scene recognition and\\global context)\end{tabular} \\ 
\midrule
\textbf{Yang} \cite{yang2022emotion} & \textbf{37.73} & \begin{tabular}[c]{@{}l@{}}5 (face landmarks, body pose, context,\\relationships, and agent-object interaction) \end{tabular} \\ 
\midrule
\textbf{Ours} & \textbf{30.02} & 1 (single-stream context) \\
\bottomrule
\end{tabular}
}
\\
\small{* Chen's reproduction of the work, as reported in \cite{chen2023incorporating}}

\label{sota}
\end{table}
\section{Results and discussion}
\label{sec:res_dis}

We compare our results with different approaches in \autoref{sota}. Our proposed method outperforms other graph-related methods, such as the work by Zhang, Liang, and Ma \cite{zhang2019context}, and also DRM and LEKG, which are two variations of Chen \textit{et al.} \cite{chen2023incorporating}. We did not compare our method with Chen's TEKG model because it has a local approach that could not be extended to real-world problems by itself. Although EmotiCon \cite{Mittal_2020_CVPR} reported a higher mAP than our method by 5.46 mAP, according to Chen \textit{et al.} \cite{chen2023incorporating}, their result of 35.48 mAP is not reproducible, reporting a score of 26.87 mAP, in which our method can perform by 3.15 mAP. Additionally, EmotiCon uses four nonverbal cues, while we only employ one. Yang \textit{et al.} \cite{yang2022emotion} reports the highest mAP in this dataset, with a result of 37.73 mAP, which is 7.71 mAP higher than our result, by also employing five nonverbal cues. Therefore, we demonstrate that our model is competitive with the state-of-the-art, even with just one cue. While Zhang, Liang, and Ma \cite{zhang2019context} also use only one cue, they process it at two levels and can be considered two contextual cues.

The number of nonverbal cues employed is directly related to the inference time of the model, a question that we wanted to tackle. Emotion recognition models should be easily deployed on edge when thinking about real-world situations. This would allow for multiple data capture and processing points without increasing spending too much or requiring high energy consumption. These are two current barriers imposed when deploying deep learning models in the wild. However, in cases where multiple cues are needed, our model could act as a context encoding stream, even with a convolutional neural network, to extract descriptions and contribute to the overall perception of emotion.

We conducted ablation studies on our model to assess its performance under different configurations. Our findings, present in \autoref{ablation}, indicate that using sum pooling instead of the current pooling approach for GIN results in inferior performance. Furthermore, we also compared the performance of our model with that of using a simple GCN consisting of two GCN blocks, ReLU activations, and a classifier. Our model outperformed this simple GCN, which resulted in a lower mAP.

\begin{table}[b!]
\caption{Ablation study of the proposed method.}
\centering
\resizebox{0.35\columnwidth}{!}{%
\renewcommand{\arraystretch}{1.1}
\begin{tabular}{ll} 
\toprule
\textbf{Method} & \textbf{mAP} \\ 
\hline\hline
GIN \cite{xu2018powerful} (AvgPooling) & 0.3002 \\
GIN \cite{xu2018powerful} (SumPooling) & 0.1715 \\
Simple GCN & 0.2505 \\
\bottomrule
\end{tabular}
}

\label{ablation}
\end{table}

\begin{figure*}[t!]
\centering
\begin{subfigure}{0.30\textwidth}
\centering
\includegraphics[width=\columnwidth]{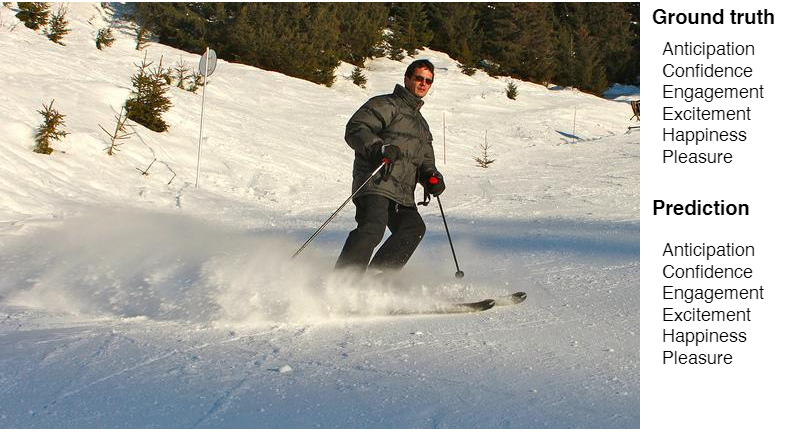}
\caption{}
\label{fig:quali:a}
\end{subfigure}
\begin{subfigure}{0.30\textwidth}
\centering
\includegraphics[width=\columnwidth]{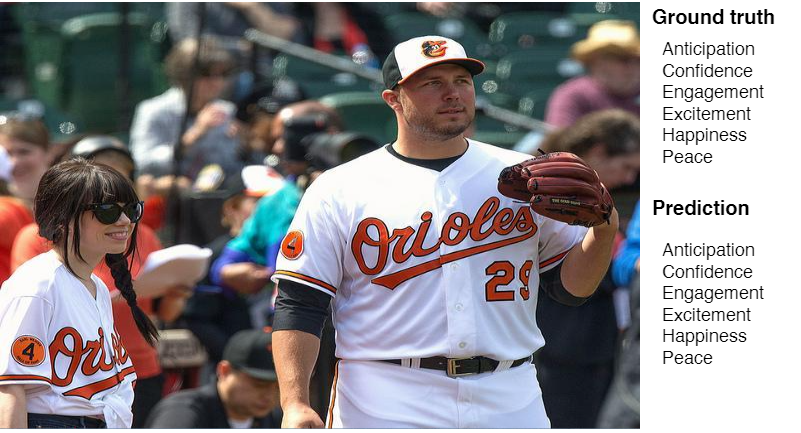}
\caption{}
\label{fig:quali:b}
\end{subfigure}
\begin{subfigure}{0.30\textwidth}
\centering
\includegraphics[width=\columnwidth]{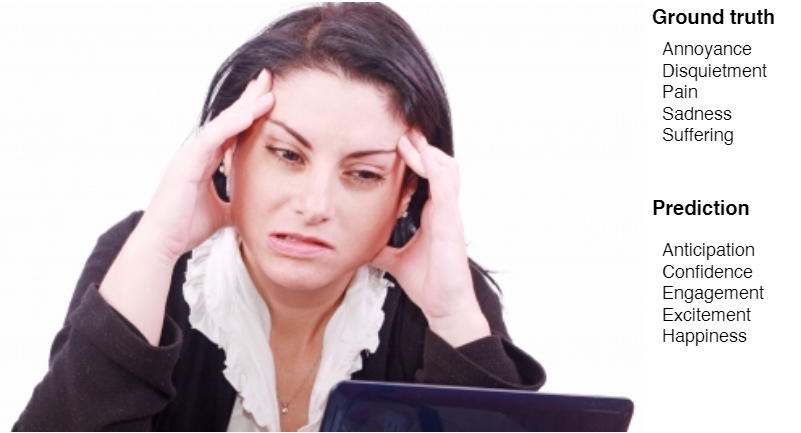}
\caption{}
\label{fig:quali:c}
\end{subfigure}

\begin{subfigure}{0.30\textwidth}
\centering
\includegraphics[width=\columnwidth]{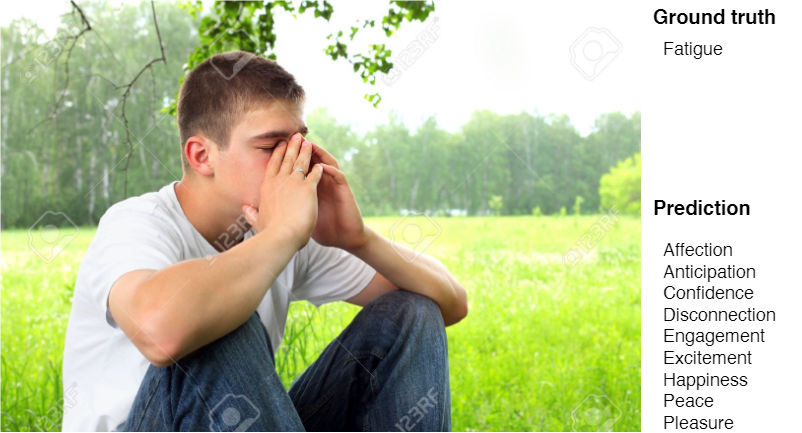}
\caption{}
\label{fig:quali:d}
\end{subfigure}
\begin{subfigure}{0.30\textwidth}
\centering
\includegraphics[width=\columnwidth]{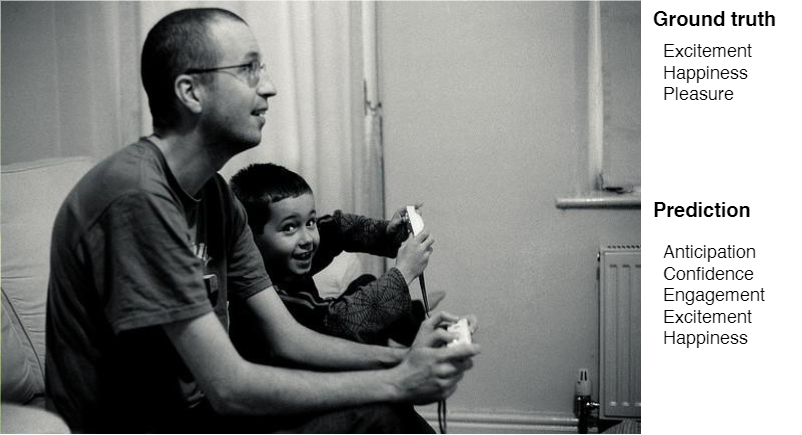}
\caption{}
\label{fig:quali:e}
\end{subfigure}
\begin{subfigure}{0.30\textwidth}
\centering
\includegraphics[width=\columnwidth]{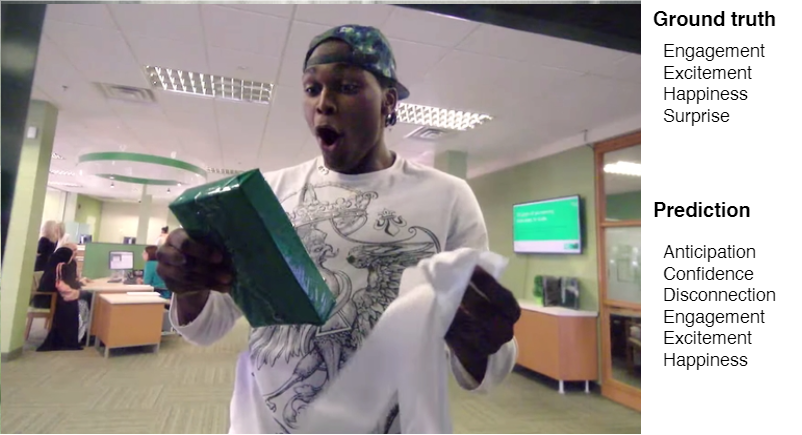}
\caption{}
\label{fig:quali:f}
\end{subfigure}

\caption{Qualitative results of our model on the EMOTIC dataset. For each image, we have the ground-truth emotion as annotated in the dataset and the prediction of the network.}
	\label{fig:qualis}
\end{figure*}

We also evaluate our model with a qualitative analysis, as shown on \autoref{fig:qualis}. For each model on the EMOTIC test set, we feed this image to the proposed pipeline, generating categorical predictions for the image. Although the model can also generate results using the VAD model, it is difficult for humans to understand and compare these values since this is unnatural for us. Therefore, we choose to use only categorical values for the qualitative evaluation. This experiment shows, as expected that our method looks for cues in context for emotion prediction. In \autoref{fig:quali:a} and \autoref{fig:quali:b}, the model could predict all categories present on the ground truth. In this case, given how the context is representative, our model can act well and give an overview of the emotion in that scene. In opposite cases, such as \autoref{fig:quali:c}, the context is not representative, and the network cannot predict any correct emotion class. In \autoref{fig:quali:d}, the context is related to a set of emotions, for example, positive emotions, but the perceived emotion of the person is actually negative. In this example, when looking at the person, we can perceive an emotion related to tiredness, which is confirmed by the ground truth \textit{Fatigue}. However, since the model does not look at face or body language from context, it perceives the wrong emotion. Finally, for \autoref{fig:quali:e} and \autoref{fig:quali:f}, the context is very generic, but the model can extract cues from it and classify correctly, at least on some level.

Finally, we test our model on different environments to assess the computational power required and the inference time. We execute the entire testing pipeline by setting a batch size of 1 for individual predictions and evaluate on both environments described in Section \ref{sec:exp}. We store each individual prediction into a list and then compute the minimum and average values. The minimum value indicates the sample in which the inference was faster, while the average value indicates the average inference time for the model. In a moderate deep learning machine, the inference of our model took 4.0264ms as a minimum, and 4.1546ms on average, leading to $\approx248$ fps and $\approx240$ fps, respectively. For a consumer-grade notebook, the inference of our model took 8.9597ms as a minimum, and 10.3898ms on average, leading to $\approx111$ fps and $\approx96$ fps, respectively. Finally, on the same consumer-grade notebook without using CUDA, the inference of the model took 13.0021ms as a minimum and 17.7365ms on average, leading to $\approx77$ fps and $\approx56$ fps, respectively, on an Intel Core i7-7700HQ @ 2.80GHz CPU.

We do not compare our inference time with the other techniques we evaluated above since neither has official open-source implementations. However, we may infer that models such as EmotiCon, which uses various other deep learning models to extract and process specific cues, would take longer than ours for execution.

\section{Conclusion}
\label{sec:concl}
In this work, we present an approach for emotion recognition based on contextual cues of the image. Our proposed approach uses image captioning to generate high-level descriptions of the image. Then, together with data mining and semantic analysis of the words, we generate and employ graphs that are high-level representations of the context. Finally, we employ graph classification approaches to classify the graphs among a set of emotional categories. Although other authors are basing their approach on multi-modality and various descriptions of context, this simple approach is effective and comparable to the more robust models. Finally, we show that our technique is fast and requires low computational power, allowing for easy and cheap deployment in multiple scenarios.

For future works, we plan on combining these high-level representations of context into a multi-cue model that also works with faces and body language to tackle the cases in which the context did not match the perceived emotion of people on the scene.

\section*{Acknowledgments}
This study was financed in part by the Coordenação de Aperfeiçoamento de Pessoal de Nível Superior – Brasil (CAPES) – Finance Code 001.

\bibliographystyle{unsrt}  
\bibliography{references}

\end{document}